\documentclass[11pt,a4paper]{article}
\usepackage[hyperref]{acl2017}
\usepackage{times}
\usepackage{latexsym}
\usepackage{graphicx}
\usepackage{url}
\aclfinalcopy

\title{Improving Distributed Representations of Tweets - Present and Future}
\author{Ganesh J \\
  Information Retrieval and Extraction Laboratory \\
  IIIT Hyderabad \\
  Telangana, India \\
  {\tt ganesh.j@research.iiit.ac.in}
}

\begin{document}
\maketitle
\begin{abstract}
Unsupervised representation learning for tweets is an important research field which helps in solving several business applications such as sentiment analysis, hashtag prediction, paraphrase detection and microblog ranking. A good tweet representation learning model must handle the idiosyncratic nature of tweets which poses several challenges such as short length, informal words, unusual grammar and misspellings. However, there is a lack of prior work which surveys the representation learning models with a focus on tweets. In this work, we organize the models based on its objective function which aids the understanding of the literature. We also provide interesting future directions, which we believe are fruitful in advancing this field by building high-quality tweet representation learning models.
\end{abstract}

\section{Introduction}
Twitter is a widely used microblogging platform, where users post and interact with messages, ``tweets''. Understanding the semantic representation of tweets can benefit a plethora of applications such as sentiment analysis~\cite{ren16_aaai, giachanou16_csur}, hashtag prediction~\cite{dhingra16_acl}, paraphrase detection~\cite{vosoughi16_sigir} and microblog ranking~\cite{huang13_cikm, shen14_cikm}. However, tweets are difficult to model as they pose several challenges such as short length, informal words, unusual grammar and misspellings. Recently, researchers are focusing on leveraging unsupervised representation learning methods based on neural networks to solve this problem. Once these representations are learned, we can use off-the-shelf predictors taking the representation as input to solve the downstream task~\cite{bengio13_slsp,bengio13_tpami}. These methods enjoy several advantages: (1) they are cheaper to train, as they work with unlabelled data, (2) they reduce the dependence on domain level experts, and (3) they are highly effective across multiple applications, in practice.

Despite this, there is a lack of prior work which surveys the tweet-specific unsupervised representation learning models. In this work, we attempt to fill this gap by investigating the models in an organized fashion. Specifically, we group the models based on the objective function it optimizes. We believe this work can aid the understanding of the existing literature. We conclude the paper by presenting interesting future research directions, which we believe are fruitful in advancing this field by building high-quality tweet representation learning models.

\begin{figure*}[tbp]
\centering
\includegraphics[width=15cm, height=8.5cm]{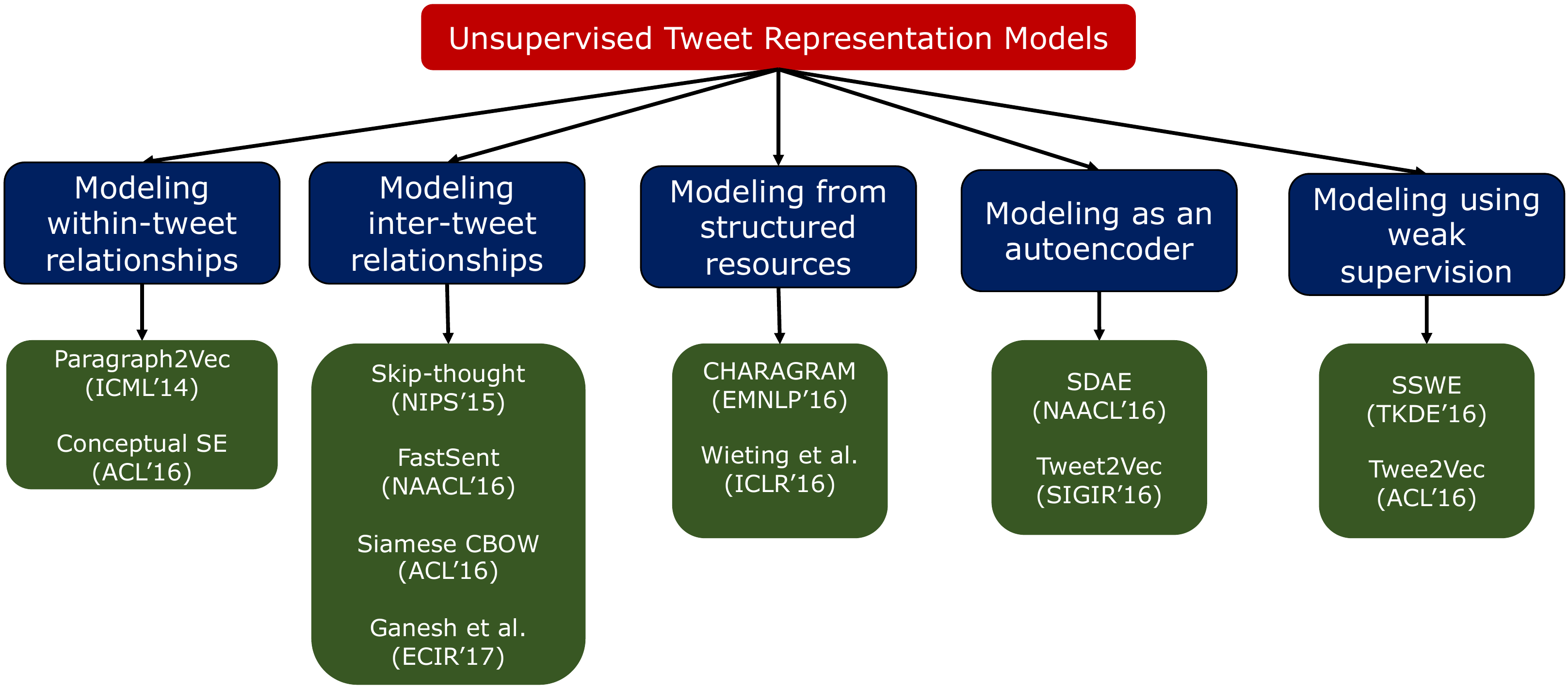}
\caption{Unsupervised Tweet Representation Models Hierarchy based on Optimized Objective Function}
\label{fig:f1}
\end{figure*}

\section{Unsupervised Tweet Representation Models}
There are various models spanning across different model architectures and objective functions in the literature to compute tweet representation in an unsupervised fashion. These models work in a semi-supervised way - the representations generated by the model is fed to an off-the-shelf predictor like Support Vector Machines (SVM) to solve a particular downstream task. These models span across a wide variety of neural network based architectures including average of word vectors, convolutional-based, recurrent-based and so on. We believe that the performance of these models is highly dependent on the objective function it optimizes -- predicting adjacent word (within-tweet relationships), adjacent tweet (inter-tweet relationships), the tweet itself (autoencoder), modeling from structured resources like paraphrase databases and weak supervision. In this section, we provide the first of its kind survey of the recent tweet-specific unsupervised models in an organized fashion to understand the literature. Specifically, we categorize each model based on the optimized objective function as shown in Figure~\ref{fig:f1}. Next, we study each category one by one. 

\subsection{Modeling within-tweet relationships}
\noindent\textbf{\underline{Motivation}}: Every tweet is assumed to have a latent topic vector, which influences the distribution of the words in the tweet. For example, though the appearance of the phrase \textit{catch the ball} is frequent in the corpus, if we know that the topic of a tweet is about ``technology'', we can expect words such as \textit{bug} or \textit{exception} after the word \textit{catch} (ignoring \textit{the}) instead of the word \textit{ball} since \textit{catch the bug/exception} is more plausible under the topic ``technology''.  On the other hand, if the topic of the tweet is about ``sports'', then we can expect \textit{ball} after \textit{catch}. These intuitions indicate that the prediction of neighboring words for a given word strongly relies on the tweet also.

\noindent\textbf{\underline{Models}}: \cite{le14_icml}'s work is the first to exploit this idea to compute distributed document representations that are good at predicting words in the document. They propose two models: PV-DM and PV-DBOW, that are extensions of Continuous Bag Of Words (CBOW) and Skip-gram model variants of the popular Word2Vec model~\cite{mikolov13_nips} respectively -- PV-DM inserts an additional document token (which can be thought of as another word) which is shared across all contexts generated from the same document;  PV-DBOW attempts to predict the sampled words from the document given the document representation. Although originally employed for paragraphs and documents, these models work better than the traditional models: BOW~\cite{harris54} and LDA~\cite{blei03_jmlr} for tweet classification and microblog retrieval tasks~\cite{wang16_acl}. The authors in \cite{wang16_acl} make the PV-DM and PV-DBOW models \textit{concept-aware} (a rich semantic signal from a tweet) by augmenting two features: attention over contextual words and conceptual tweet embedding, which jointly exploit concept-level senses of tweets to compute better representations. Both the discussed works have the following characteristics: (1) they use a shallow architecture, which enables fast training, (2) computing representations for test tweets requires computing gradients, which is time-consuming for real-time Twitter applications, and (3) most importantly, they fail to exploit textual information from related tweets that can bear salient semantic signals.

\subsection{Modeling inter-tweet relationships}
\noindent\textbf{\underline{Motivation}}: To capture rich tweet semantics, researchers are attempting to exploit a type of \textit{sentence-level Distributional Hypothesis}~\cite{harris54,polajnar2015_lsdsem}. The idea is to infer the tweet representation from the content of adjacent tweets in a related stream like users' Twitter timeline, topical, retweet and conversational stream. This approach significantly alleviates the context insufficiency problem caused due to the ambiguous and short nature of tweets \cite{ren16_aaai,ganesh17_ecir}.

\noindent\textbf{\underline{Models}}: Skip-thought vectors~\cite{kiros15_nips} (STV) is a widely popular sentence encoder, which is trained to predict adjacent sentences in the book corpus~\cite{zhu15_iccv}. Although the testing is cheap as it involves a cheap forward propagation of the test sentence, STV is very slow to train thanks to its complicated model architecture. To combat this computational inefficiency, FastSent~\cite{hill16_naacl} propose a simple additive (log-linear) sentence model, which predicts adjacent sentences (represented as BOW) taking the BOW representation of some sentence in context. This model can exploit the same signal, but at a much lower computational expense. Parallel to this work, Siamase CBOW~\cite{kenter16_acl} develop a model which directly compares the BOW representation of two sentence to bring the embeddings of a sentence closer to its adjacent sentence, away from a randomly occurring sentence in the corpus. For FastSent and Siamese CBOW, the test sentence representation is a simple average of word vectors obtained after training. Both of these models are general purpose sentence representation models trained on book corpus, yet give a competitive performance over previous models on the tweet semantic similarity computation task. \cite{ganesh17_ecir}'s model attempt to exploit these signals directly from Twitter. With the help of attention technique and learned user representation, this log-linear model is able to capture salient semantic information from chronologically adjacent tweets of a target tweet in users' Twitter timeline.

\subsection{Modeling from structured resources}
\noindent\textbf{\underline{Motivation}}: In recent times, building  representation models based on supervision from richly structured resources such as Paraphrase Database (PPDB)~\cite{ganitkevitch13_naacl} (containing noisy phrase pairs) has yielded high quality sentence representations. These methods work by maximizing the similarity of the sentences in the learned semantic space.

\noindent\textbf{\underline{Models}}: \textit{CHARAGRAM}~\cite{wieting16_emnlp} embeds textual sequences by learning a character-based compositional model that involves addition of the vectors of its character n-grams followed by an elementwise nonlinearity. This simpler architecture trained on PPDB is able to beat models with complex architectures like CNN, LSTM on SemEval 2015 Twitter textual similarity task by a large margin. This result emphasizes the importance of character-level models that address differences due to spelling variation and word choice. The authors in their subsequent work~\cite{wieting16_iclr} conduct a comprehensive analysis of models spanning the range of complexity from word averaging to LSTMs for its ability to do transfer and supervised learning after optimizing a margin based loss on PPDB. For transfer learning, they find models based on word averaging perform well on both the in-domain and out-of-domain textual similarity tasks, beating LSTM model by a large margin. On the other hand, the word averaging models perform well for both sentence similarity and textual entailment tasks, outperforming the LSTM. However, for sentiment classification task, they find LSTM (trained on PPDB) to beat the averaging models to establish a new state of the art. The above results suggest that structured resources play a vital role in computing general-purpose embeddings useful in downstream applications.

\subsection{Modeling as an autoencoder}
\noindent\textbf{\underline{Motivation}}: The \textit{autoencoder} based approach learns latent (or compressed) representation by reconstructing its own input. Since textual data like tweets contain discrete input signals, \textit{sequence-to-sequence models}~\cite{SutskeverVL14} like STV can be used to build the solution. The encoder model which encodes the input tweet can typically be a CNN~\cite{kim14_emnlp}, recurrent models like RNN, GRU, LSTM~\cite{karpathy15_iclr} or memory networks~\cite{sukhbaatar15_nips}. The decoder model which generates the output tweet can typically be a recurrent model that predicts a output token at every time step.

\noindent\textbf{\underline{Models}}: Sequential Denoising Autoencoders (SDAE)~\cite{hill16_naacl} is a LSTM-based sequence-to-sequence model, which is trained to recover the original data from the corrupted version. SDAE produces robust representations by learning to represent the data in terms of features that explain its important factors of variation. Tweet2Vec~\cite{vosoughi16_sigir} is a recent model which uses a character-level CNN-LSTM encoder-decoder architecture trained to construct the input tweet directly. This model outperforms competitive models that work on word-level like PV-DM, PV-DBOW on semantic similarity computation and sentiment classification tasks, thereby showing that the character-level nature of Tweet2Vec is best-suited to deal with the noise and idiosyncrasies of tweets. Tweet2Vec controls the generalization error by using a data augmentation technique, wherein tweets are replicated and some of the words in the replicated tweets are replaced with their synonyms. Both SDAE and Tweet2Vec has the advantage that they don't need a coherent inter-sentence narrative (like STV), which is hard to obtain in Twitter.

\subsection{Modeling using weak supervision}
\noindent\textbf{\underline{Motivation}}: In a weakly supervised setup, we create labels for a tweet automatically and predict them to learn potentially sophisticated models than those obtained by unsupervised learning alone. Examples of labels include sentiment of the overall tweet, words like hashtag present in the tweet and so on. This technique can create a \textit{huge} labeled dataset especially for building data-hungry, sophisticated deep learning models.

\noindent\textbf{\underline{Models}}:
\cite{tang16_tkde} learns sentiment-specific word embedding (SSWE), which encodes the polarity information in the word representations so that words with contrasting polarities and similar syntactic context (like \textit{good} and \textit{bad}) are pushed away from each other in the semantic space that it learns. SSWE utilizes the massive distant-supervised tweets collected by positive and negative emoticons to build a powerful tweet representation, which are shown to be useful in tasks such as sentiment classification and word similarity computation in sentiment lexicon. \cite{dhingra16_acl} observes that \textit{hashtags} in tweets can be considered as topics and hence tweets with similar hashtags must come closer to each other. Their model predicts the hashtags by using a Bi-GRU layer to embed the tweets from its characters. Due to subword modeling, such character-level models can approximate the representations for rare words and new words (words not seen during training) in the test tweets really well. This model outperforms the word-level baselines for hashtag prediction task, thereby concluding that exploring character-level models for tweets is a worthy research direction to pursue. Both these works fail to study the model's generality~\cite{weston14_emnlp}, i.e., the ability of the model to transfer the learned representations to diverse tasks.   

\section{Future Directions}
In this section we present the future research directions which we believe can be worth pursuing to generate high quality tweet embeddings.
\begin{itemize}
\item \cite{ren16_aaai} propose a supervised neural network utilizing contextualized features from conversation, author and topic based context about a target tweet to perform well in classification of tweet. Apart from \cite{ganesh17_ecir}'s work which utilizes author context, there is no other work which builds unsupervised tweet representation model on Twitter-specific contexts such as conversation and topical streams. We believe such a solution directly exploits semantic signals (or nuances) from Twitter, unlike STV or Siamese CBOW which are trained on books corpus.  
\item \cite{santos14_coling} propose a supervised, hybrid model exploiting both the character and word level information for Twitter sentiment analysis task. Since the settings when the character level model beats the word level model is not well understood yet, we believe it would be interesting to explore such a hybrid compositional model to build unsupervised tweet representations.
\item Twitter provides a platform for the users to interact with other users. To the best of our knowledge, there is no related work that computes unsupervised tweet representation by exploiting the user profile attributes like profile picture, user biography and set of followers, and social interactions like retweet context (set of surrounding tweets in a users’ retweet stream) and favorite context (set of surrounding tweets in a users’ favorite tweet stream). 
\item DSSM~\cite{huang13_cikm,shen14_cikm} propose a family of deep models that are trained to maximize the relevance of clicked documents given a query. Such a ranking loss function helps the model cater to a wide variety of applications\footnote{\url{https://www.microsoft.com/en-us/research/project/dssm/}} such as web search ranking, ad selection/relevance, question answering, knowledge inference and machine translation. We observe such a loss function has not been explored for building unsupervised tweet representations. We believe employing a ranking loss directly on tweets using a large scale microblog dataset\footnote{\url{http://trec.nist.gov/}} can result in representations which can be useful to Twitter applications beyond those studied in the tweet representation learning literature.
\item Linguists assume that language is best understood as a hierarchical tree of phrases, rather than a flat sequence of words or characters. It's difficult to get the syntactic trees for tweets as most of them are not grammatically correct. The average of word vectors model has the most simplest compositional architecture with no additional parameters, yet displays a strong performance outperforming complex architectures such as CNN, LSTM and so on for several downstream applications~\cite{wieting16_emnlp,wieting16_iclr}. We believe a theoretical understanding of why word averaging models perform well can help in embracing these models by linguists.
\item Models in~\cite{wieting16_emnlp,wieting16_iclr} learn  from noisy phrase pairs of PPDB. Note that the source of the underlying texts is completely different from Twitter. It can be interesting to see the effectiveness of such models when directly trained on structural resources from Twitter like Twitter Paraphrase Corpus~\cite{xu14_tacl}. The main challenge with this approach is the small size of the annotated Twitter resources, which can encourage models like \cite{sanjeev17_iclr} that work well even when the training data is scarce or nonexistent.  
\item Tweets mostly have an accompanying image which sometimes has visual correspondence with its textual content~\cite{chen13_mm,wang14_tomccap} (`visual' tweet). To the best of our knowledge, there is no work which explores the following question: \textit{can we build multimodal representations for tweets accompanying correlated visual content and compare with traditional benchmarks?}. We can leverage insights from multimodal skip-gram model~\cite{lazaridou15_naacl} which builds multimodally-enhanced word vectors that perform well in the traditional semantic benchmarks. However, it's hard to detect visual tweets and learning from a non-visual tweet can degrade its tweet representation. It would be interesting to see if a dispersion metric~\cite{kiela14_acl} for tweets can be explored to overcome this problem of building a nondegradable, improved tweet representation.
\item Interpreting the tweet representations to unearth the encoded features responsible for its performance on a downstream task is an important, but a less studied research area. \cite{ganesh17_nips}'s work is the first to open the blackbox of vector embeddings for tweets. They propose elementary property prediction tasks which predicts the accuracy to which a given tweet representation encodes the elementary property (like slang words, hashtags, mentions, etc). The main drawback of the work is that they fail to correlate their study with downstream applications. We believe performing such a correlation study can clearly highlight the set of elementary features behind the performance of a particular representation model over other for a given downstream task.   
\end{itemize}

\section{Conclusion}
In this work we study the problem of learning unsupervised tweet representations. We believe our survey of the existing works based on the objective function can give vital perspectives to researchers and aid their understanding of the field. We also believe the future research directions studied in this work can help in breaking the barriers in building high quality, general purpose tweet representation models.

\bibliography{acl2017}
\bibliographystyle{acl_natbib}

\end{document}